\documentclass[pdflatex,sn-mathphys]{sn-jnl}

\jyear{2021}%

\theoremstyle{thmstyleone}%
\theoremstyle{thmstyletwo}%

\theoremstyle{thmstylethree}%

\begin{document}
\title[Sah et al.]{\textbf{Design of Low Thrust Controlled Maneuvers to Chase and De-orbit the Space Debris}}

\author*[1]{\fnm{Roshan} \sur{Sah}}\email{sah.roshan@tcs.com}
\author [2]{\fnm{Raunak} \sur{Srivastava}}\email{srivastava.raunak@tcs.com}
\equalcont{These authors contributed equally to this work.}
\author [3]{\fnm{Kaushik} \sur{Das}}\email{kaushik.da@tcs.com}
\equalcont{These authors contributed equally to this work.}
\affil[1]{\orgdiv{TCS Research}, \orgname{Tata Consultancy Services}, \orgaddress{\city{Bangalore}, \postcode{560066}, \state{Karnataka}, \country{India}}}


\abstract{Over the several decades, the space debris at LEO has grown rapidly which had caused a serious threat to the operating satellite in an orbit. To avoid the risk of collision and protect the LEO environment, the space robotics ADR concept has been continuously developed over a decade to chase, capture and de-orbit space debris. This paper presents the designed small satellite with dual robotic manipulators. The small satellite is designed based on CubeSat standards, which uses commercially available products in the market. In this paper, an approach is detailed for designing the controlled chase and de-orbit maneuver for a small satellite equipped with an RCS thruster. The maneuvers are comprised of two phases: a)bringing the chaser satellite to the debris orbit, and accelerating it to close proximity of 1m to debris object by using the low thrust RCS thruster, and b) Once captured, controlled de-orbiting it to 250 km of altitude. A Hohmann transfer concept is used to move our chaser satellite from the lower orbit to the debris orbit by two impulsive burns. A number of the scenarios are simulated, where one or more orbital elements are adjusted. For more than one orbital elements adjustment, the Directional Adaptive Guidance law (DAG- law) and the Proximity Quotient Guidance laws (Q- law) are utilized. These laws synthesize the three direction thrusts to the single thrust force for the controlled maneuver. The amount of propellant consumed and the thrust characteristics at each maneuver are determined by using the performance parameters of the RCS thruster intended for a small satellite. The results show that, for long-term simulation of chaser satellite’s maneuver to debris object, an optimum DAG law is most suitable than the Q- law, as it can handle the singularity behaviour of the orbital elements caused due to adjustment of one or more elements more efficiently. Numerous scenarios are simulated to investigate the feasibility of the controlled chase close proximity, and de-orbiting operation to debris objects by using RCS thruster and optimal control law.}



\maketitle
\vspace{-10mm}
\section{Introduction}
\label{Introduction}

The low earth orbit(LEO) space environment is continuously crowded with space debris. This debris mostly consists of the residual of the spent satellite, rocket staging, body and booster, junk particles from the collision of debris object. Around mid of the twentieth century, space junk was not considered a serious issue as fewer applications were known at LEO. Recently, the space junk has increased drastically in the various belt of existing satellite orbits. Rex et al. study show that the debris junk particle will grow with a 5\% growth rate per year if the possible debris mitigating measures are not taken\cite{rex_eichler_1993}. The post-mission removal of the spent satellites and the rockets bodies has become important for keeping the functional satellites in favorable orbital condition and for upcoming space missions. Most of the studies shows that spacecraft's uncontrolled space debris collisions had increased continuously in orbital belts\cite{nasa_2000} The accumulation of the space junk in existing orbits can cause the possibility of the collision, which can cause an increment of more debris into the orbital belt. The augmented debris can be grouped as the "Kessler Syndrome" which could have a major impact worldwide and the frequency of the collisions will increase if post-mission removal is not done \cite{Kessler10thekessler}. The orbiting debris is capable of having a relative speed of 15000 mph and can cause serious damage to the existing satellites in the orbit belts. Pelton describes how the debris generates by the cascading effect caused due to collision of space debris. He also explains the international standards for space traffic management and mitigating debris for future stratospheric missions and activities \cite{pelton}. According to the European Space Agency(ESA), it has become mandatory to de-orbit the satellites bodies within the twenty-five years of its end of life (EOD). Moreover, de-orbiting becomes an essential process to remove the debris object from the orbital belts to reduce future collision probability. In the future, it is necessary to conserve the debris environment with minimized risk, mainly in the LEO region\cite{klinkrad1999esa}. Anselmo et al. article suggest that the long-term evolution of debris population collision risk can be reduced by the explosion avoidance strategies and de-orbiting of upper stages in LEO regions\cite{ANSELMO20011427}.

To reduce the debris growth in LEO orbit, there are typically two measures defined, one is debris avoidance and the other is debris removal. In debris avoidance, the operation spacecraft or satellites use in-flight maneuvers to avoid collision with the space junk. But in debris removal, space junk from the orbit is removed by the means of other spacecraft and de-orbit to very low earth orbit or transferred to the graveyard orbit. 
Basically, there are two debris removal approaches, they are

1. Active Debris Removal (ADR)

2. Passive Debris Removal (PDR)

ADR concept has been continuously developed for over a decade for the removal of large space debris object.
Whereas, PDR approaches are used for the removal of the small debris. As the small debris has enough kinetic energy which is capable of destroying the operational satellites. 
As the ADR concept mostly uses the propulsion system for its operation. Over the last couple of decades, the advancement in the propulsion system has raised continuously for the number of missions. Previously, the solid and liquid propulsion systems are continuously used for different satellite missions, interplanetary mission, etc. for orbit transfer, maneuvers, docking,etc. operation. But the continuous up-gradation and miniaturization in these propulsion systems have caused the generation of the hybrid propulsion system like the RCS thruster. This kind of thruster can be used for both maneuvering and controlling the orbit transfer and chase to the close proximity of the target satellite. Various space agencies like NASA, ESA, and universities like UTIAS SFL, Surrey space center, JPL Caltech, DLR (German Aerospace Center) Braunschweig, University of Patras, etc. are working in chasing, capturing, and de-orbiting a spent satellite. Recently, Astro-scale  Japan had successfully demonstrated chase, in-space capture, and the release system to clean the space debris \cite{doi:10.1080/01691864.2021.1976671},\cite{fujii}.

Recently active debris removal has continuously evolved for the mitigation of the debris from the LEO space environment. The ADR concept consists of different methods like space robotics, tether-based , collective, laser-based, sail-based, ion beam, dynamic system-based,etc.\cite{MARK2019194} One of the recently evolving ADR methods is space robotics for debris removal.  

Over the last decade, space-based manipulators is continuously evolving approaches for the multiple on-orbit servicing (OOS) missions like debris removal, refueling, docking, assembly functions, transporting, berthing, etc. But the debris removal has become an emergent application for keeping a safe space environment. A number of the OOS mission were  completed successfully for different application except for debris removal. Space robotics has been installed in the international space station (ISS) for the assembly and servicing purpose which consists of three manipulator systems; European Robotic Arm \cite{BOUMANS199817}, Canadian Mobile Services System, and Japan Experimental Modules Manipulator System.
Recently, a satellite equipped with robotic manipulator systems with grasping devices on it for the active debris removal application has been comprehensively developed.

This paper presents the designed satellite with dual manipulators known as debris chaser satellite \cite{roshan}\cite{raunak} for the debris removal application. The satellite is designed based on the CubeSat standard Which is 12U and commercially available products in the market. The paper technically presents a simulation approach in designing the controlled chase and de-orbiting maneuver of the debris chaser satellite for the PSLV debris removal. The maneuvers are comprised of two phases: 

a)bringing the chaser satellite to the debris orbit, and accelerating it to close proximity of 1m in-track  separation to real PSLV Debris through the impulsive maneuvers, and 

b) Once captured, controlled de-orbiting it to 250 km of altitude.

Once, the PSLV debris is in a very low earth orbit, atmospheric drag and solar radiation activity projection on it will be enough to burn the structure and decay. The Directional Adaptive Guidance law (DAG- law) and the Proximity Quotient Guidance laws (Q- law)  are explained in the paper and their importance was analyzed for our simulation cases. And the DAG  guidance law is utilized to execute our controlled chase maneuver for our scenarios by using the RCS thruster at optimum thrust requirement.

\section{Debris Chaser Mission and Architecture}
\label{Debris Chaser Mission and Architecture}

The debris chaser mission consists of debris assessment, chase maneuver operation, robotic manipulator deployment, and de-orbiting operation. These mission operations are shown in figure 1 in the block diagram format.

\begin{figure}[ht]
    \centering
    \includegraphics[width =0.98\textwidth,height =3.0 in]{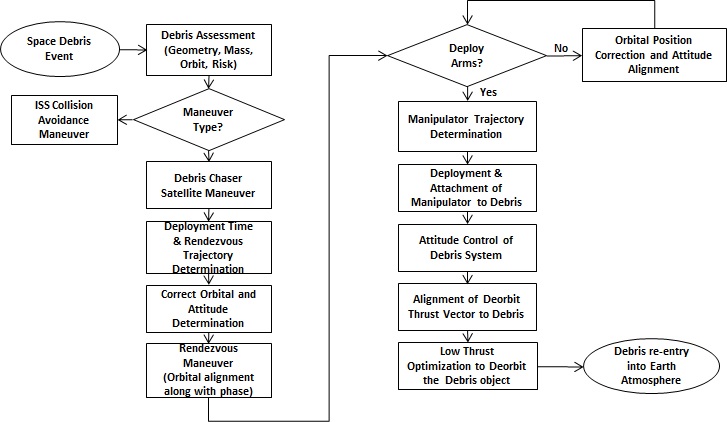}
    \centering
    \caption{Debris chaser mission diagram.}
    \centering
    \label{fig:1}
\end{figure}

\vspace{-5mm}
All the simulation platform is developed based on the above mission block diagram.
Initially, the investigation of the selected debris properties like orbit location, size, shape, orbital speed, etc. is to be done through the debris assessment tools. Then allocation of the debris chaser is done at optimal orbit height based on the capability of its maneuvers. The orbital maneuver like orbit transfer and chase maneuver is done to reach close to the debris targets and the orbital alignment along the phase is executed. Once reaching nearby, all the orbital elements and attitude orientation is done by attitude determination and control system of the debris chaser satellite
After that, on-orbit deployment of robotics hands will be done by the manipulator trajectory which will be determined by using the tracking camera which is installed in the robotic arms. The robotic arm will be deployed and debris objects get attached to our satellite. Then reaction control system (RCS) propulsion system will be used for controlling the disoriented motion of debris along with the debris chaser satellite. Then de-orbiting technique will be employed for moving the satellite to very low earth orbit. The system will be de-orbited and release the debris object towards 250 km altitude and the atmospheric drag would be enough to bring back to earth atmosphere and burn it.

The designed debris chaser satellites \cite{roshan} are shown in figure 2 and figure 3. The designed satellite is of 12U form factor with an approximate dimension of 20X20X30 cm with dual robotic manipulators. Figure 2 represents the chaser satellite in the stowed configuration whereas figure 3 represents a satellite with a robotic manipulator with an extended arm.

  \begin{figure}[ht]
   \centering
     \begin{minipage}{.49\textwidth}
      \centering
      \includegraphics[width =0.98\textwidth,height =1.8in]{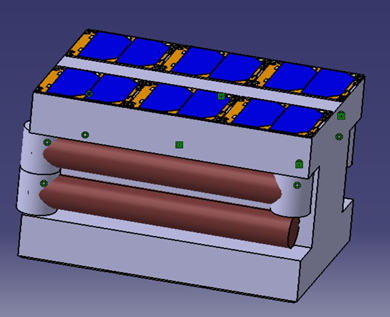}
      \centering
      \caption{ Chaser satellite in stowed configuration.}
       \label{fig:2}
       \end{minipage}
    \begin{minipage}{.49\textwidth}
    \centering
      \includegraphics[width =0.98\textwidth,height =1.8in]{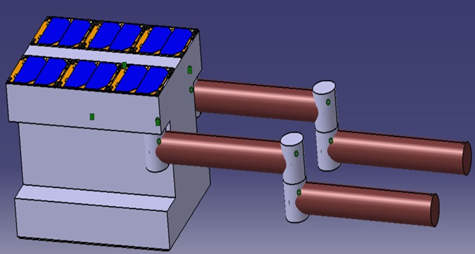}
     \centering
      \caption{Chaser satellite with robotic arm extended.}
      \label{fig:3}
     \end{minipage}
\end{figure}

\vspace{-5mm}
All the cad design is executed in the CATIA V5 tool which gives us iterative design with a higher accuracy level. A customized 3-DOF manipulator is designed based on the concept of the UR5 robot and resizing is done, so that manipulators got adjusted with the satellites model. The design of the hand gripper is still in progress which will be designed based on the grasp points in the debris.

Similarly, the system architectures \cite{Roshan2021} were designed based on the optimal requirement of the hardware and the software for successful removal of the debris at polar orbit. The schematic diagram of the system architecture is shown in figure 4. The blue and red lines represent the data and power input/output section. From figure 4, we can also see that, for the application of the debris chaser, ADCS \cite{Raunak2021} requires huge amounts of sensors which will help our debris chaser satellite to capture the debris and de-orbit it.

\begin{figure}[ht]
    \centering
    \includegraphics[width =0.9\textwidth,height =2.5 in]{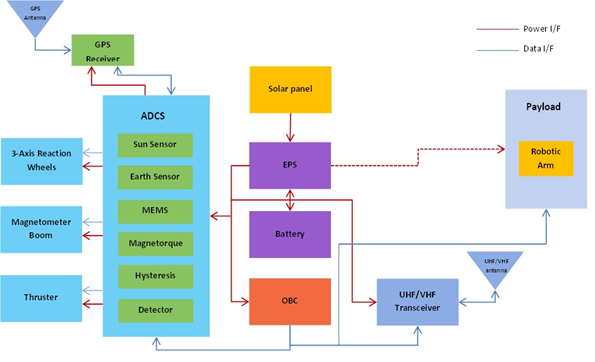}
    \centering
    \caption{Debris chaser systems architecture diagram.}
    \label{fig:4}
\end{figure}

Similarly, the thruster will play an important role in debris removal operations as thrust is required while executing the transfer, chase, and de-orbiting operations. Numerous literature surveys were done for all different types of the propulsion system  \cite{doi:10.1063/1.5007734}\cite{RANJAN2017754}, where we found that, for our application scenario, the RCS system is suitable for our transfer and rendezvous maneuvers.

For our debris chaser satellite, we will be using the 3 RCS thruster which will be oriented in all three directions.  The RCS thruster is made of the reaction control wheel and thruster whose main function is to provide the required thrust in any direction and control the attitude motion of the debris chaser satellite. Sometimes, this kind of thruster helps in providing the torque to control the rotational motion like pitch, roll, and yaw. we are using the commercially available product of the RCS system. A  survey for the different propulsion systems from the existing literature was conducted and a feasible option was found based on the compatibility and available product in the market. It was found that the VACCO chemically-etched micro propulsion system modules\cite{Cardin2000}\cite{Cardin2003}, specifically the hybrid ADN delta-V/RCS system is more compatible and feasible for our debris chaser mission. It is the single axial high-thrust with a high specific impulse ADN thruster which can deliver up to 1,036 Ns total impulse using only integral propellant. After doing the trade-off analysis, we found that the VACCO modules can achieve the delta-V required within two hours. Here, we assumed that the thruster is firing constantly which may not be a realistic case; however, the magnitude is acceptable.

\section{Orbital Maneuver}
\label{Orbital Maneuver}
An orbital maneuver is one of the important parts of astrodynamics which transfer the spacecraft from one orbit to another orbit. It requires the propulsive thrust to do so. It is one of the most effective methods for transfer, chase, docking, and de-orbiting operation for the debris removal application. For the orbit transfer, Hohmann's method is commonly used and is an effective technique while executing the transfer. It is considered as a simplest and most efficient method of transferring a satellite in the co-apsidal axis and co-planar orbits. It is a two-impulse elliptical transfer between two co-planar circular orbits. The transfer itself consists of an elliptical orbit with a perigee at the inner orbit and an apogee at the outer orbit. Figure 5 shows Hohmann's transfer orbit with the direction of net velocity after firing the propulsion unit.

\begin{figure}[ht]
    \centering
    \includegraphics[width =0.4\textwidth,height =2in]{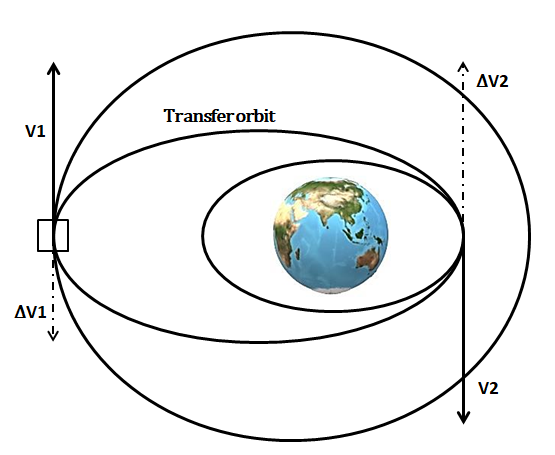}
    \centering
    \caption{Hohmann's Transfer Orbit.}
    \label{fig:5}
\end{figure}

 \vspace{-5mm}
 The governing equations of Hohmann's orbit transfer can be found widely in literature, which is one of the basic methods for the execution of orbit transfer to a higher orbit.
 
For the in-orbit transfer to the debris object, a rendezvous chase maneuver like R-bar, V-bar, and Z-bar traditional approach are most commonly used. In the R-bar approach, a satellite chase from below or above the target object, along its radial direction. In the V-bar approach, a satellite chase from ahead or from behind and in the same direction as the orbital motion of the target object. The satellite motion is parallel to the target's orbital velocity. In it, a thruster fires small amount of fuel to increase its velocity in the direction of the target while chasing form behind. For our case, we will be using the V-bar approach to chase the space debris from behind by using an RCS thruster and will reach close proximity to the debris object. 
Once reaching the nearby, robotic manipulators will be used for grasping the non-cooperative debris through the contact dynamics method (which is not presented here).
After grasping and ceasing the motion o of the debris object, the debris chaser satellite with debris will become a single body and need to de-orbit to very low earth orbit.
For de-orbiting operation, the same Hohmann's orbit transfer will be used for the execution and moving debris to approximate 250 km where an atmospheric drag and solar projection towards debris will be enough to burn and decay it.

\section{Guidance Algorithm}
To optimize the orbital trajectory during rendezvous and close proximity operation, low thrust maneuver techniques are highly considered during its operation. Numerous researchers had considered different optimal low thrust algorithms for the rendezvous and docking process which are still in use.  An optimal steering algorithm has been developed by Edelbaum T. \cite{doi:10.2514/8.5723} which is used for continuous thrusting in circular orbit for a given classical orbital element. For low thrust and controlled maneuvers, two guidance laws are continuously used for the adjustment of one or multiple orbital elements. They are; one is Proximity Quotient (Q) law \cite{doi:10.2514/6.2004-5089} and another is Directional Adaptive Guidance (DAG) law \cite{doi:10.2514/6.2014-3714}.

\subsection{ Proximity Quotient (Q) law}

The proximity Quotient law is also known as the Q law. It was first developed by the Petropoulos\cite{doi:10.2514/6.2004-5089}\cite{petro} in 2003 to find the initial assumption for optimal propellant and low thrust transfer between 2 Keplerian orbits. The Q-law is generally based on the Lyapunov feedback control loop, which calculates the optimal thrust direction i.e. $\alpha$ and $\beta$, based on the initial orbit of the spacecraft and desired target orbit. The  Q-law was mostly developed for the 2 body problem which is based on the Gauss planetary equations and helps us to determine the rate of the Layapunov function analytically. 

Based on Layapunov function, the Q is defined as:

\vspace{-5mm}

\begin{equation}
\boldsymbol
  Q= (1+ W_{p}P) \sum_{COE} W_{COE}S_{COE} \Bigg[\frac{COE-COE_{t}}{\dot{(COE)_{xx}}}\Bigg]^2
\end{equation}

Where, P represents the penalty introduced in the algorithm to find the solution, otherwise it can be involved in unacceptable low periapsis altitudes, $W_{p}$ is the weighting factor in the P term. The $S_{COE}$ is introduced to scale down the semi-major axis error and to improve the convergence in the simulation and $W_{COE}$ is the weighting factor that has a significant role in changing each COE quickly. The term  $\dot{(COE)_{xx}}$  is the maximum rate of change of a given COE within the instantaneous orbit. 
The expression and governing equation of the  $S_{COE}$ and P is mentioned below;
\begin{equation}
\boldsymbol
  S_{a} = \sqrt{\Bigg[1+\Bigg(\frac{\Delta(a,a_{t})}{3a_{t}}\Bigg)^4}\Bigg]
\end{equation}

\begin{equation}
\boldsymbol
  S_{e} = S_{i}=S_{\Omega}=S_{\omega}=1
\end{equation}

Where $a$ and $a_{t}$  are the instantaneous and the target values of the semi-major axis.

\begin{equation}
\boldsymbol
  P= exp \Bigg[k\Bigg(1-\frac{r_{p}}{r_{pmin}}\Bigg)\Bigg]
\end{equation}

Where, K  denotes the penalty function strength, $r_{p}$ and $r_{pmin}$ are the instantaneous and the minimum allowable periapsis radius.

The important idea behind the proximity quotient is that the Q usually represents some error for the desired state. The expression of $\dot{Q}$ can be determined analytically by  using the Gauss form of the Lagrange Planetary equations and  will involve the unit thrust angle $\alpha$ and $\beta$ (i.e. pitch and yaw angle.) are shown below;

\begin{equation}
\boldsymbol
  {\dot{Q}} = \sum_{COE} \frac{\partial{Q}}{\partial{COE}}*{\dot{COE}}
\end{equation}
From the above equation, an analytical solution exists which minimizes the  $\dot{Q}$  by varying the thrust angle $\alpha$ and $\beta$.  By using these angles of the satellite derives Q to zero at a given instant and thus driving the satellite to the desired target orbit.

\subsection{Directional Adaptive Guidance (DAG) Law}

The optimal guidance for each orbital element is obtained in the adaptive method then clustered them by weighted-sum approach, which will give us a single thrust vector for the rendezvous, proximity, and de-orbiting operation. Factors like thrust angle expressions, maneuver efficiency, adaptive weighting factors are also considered in a single thrust vector. This law is implemented as a self-contained routine requiring the vehicle state vector and epoch of interest along with desired final targets, directional weighting factors, and stopping tolerance with the control rate limit. These implementations are mostly represented in the RIC frame. RIC frame represents R = Radial direction, I= In track direction, and C = Cross track, whose origin is the center of the satellite. The RIC frame is the relative motion frame between the chaser satellite and target debris.

The thrust angling of the vehicle state vector at an instant was computed; They are represented as; $\alpha$ is the in-plane angle (Pitch) and $\beta$ is the out-plane angle (Yaw). These angles mostly produce the largest rate of change in all the orbital element parameters like semi-major axis (a), inclination (i), eccentricity (e), right ascension of the ascending node ($\Omega$), and argument of Perigee ($\omega$). The optimal results obtained from the classical orbital elements in-plane and out-plane are found from the literature \cite{Ruggiero2011LowThrustMF} which is given below;

 Semi-major axis (a);
\begin{equation}
\boldsymbol
  \alpha = \arctan(\frac{e\sin{\theta}}{1+e\cos{\theta}}),  \beta = 0
\end{equation}

Where, $\theta$ represent the true anomaly in degree.

   eccentricity (e);
\begin{equation}
\boldsymbol
  \alpha = \arctan(\frac{\sin{\theta}}{\cos{\theta +\cos{E}}}),  \beta = 0
\end{equation}

Where, E represent the eccentricity anomaly.

 Inclination (i);
\begin{equation}
\boldsymbol
  \alpha = 0,  \beta = sgn(\cos{(\omega+\theta)})*\frac{\Pi}{2}
\end{equation}

Ascending Node ($\Omega$);

\begin{equation}
\boldsymbol
  \alpha = 0,  \beta = sgn(\cos{(\omega+\theta)})*\frac{\Pi}{2}
\end{equation}

Argument of Periapsis ($\omega$);

\begin{equation*}
\boldsymbol
  \alpha = \arctan(\frac{1+e\cos{\theta}}{2+e\cos{\theta}}*\cot{\theta})
\end{equation*}

\vspace{-3mm}

\begin{equation}
\boldsymbol
  \beta = \arctan(\frac{e\cot{(i)}\sin{(\omega+\theta)}}{\sin{(\alpha-\theta)}(1+e\cos{\theta})-\cos{\alpha}\sin{\theta}})
\end{equation}

The optimal thrust angles i.e, in-plane($\alpha$) and out-of-plane ($\beta$) for the maximum instantaneous change in each classical orbital element (COE)  can be represented by the above solutions. 
After having all the thrusting angles from the above equations[1-5], the corresponding thrust direction can be computed in the RIC frame which is given below;

\begin{equation}
\boldsymbol
{\vec{f}_{COE}} = \begin{bmatrix} f_{R} \\ f_{S} \\ f_{W} \end{bmatrix} = \begin{bmatrix} \cos{(\beta)}\sin{(\alpha)} \\ \cos{(\beta)}\cos{(\alpha)} \\ \sin{(\beta)} \end{bmatrix}
\end{equation}

For our desired target of PSLV debris, an adaptive ratio is computed which signifies the percentage change that occurred in each COE. The computed adaptive ratio can be shown as;

\begin{equation}
\boldsymbol
R_{COE} = \frac{COE_{t}-COE}{COE_{t}-COE_{i}}
\end{equation}

Where COE represents the classical orbital element at a certain instant, $COE_{t}$ indicates the COE of the desired target, and the $COE_{i}$ is the initial value of COE of the chaser satellite.

Once the adaptive ratio is calculated,  the overall unit thrust direction to achieve our target at multiple-element change at the same time can be calculated by the below formula which can be found in the literature\cite{doi:10.2514/6.2014-3714}\cite{Ruggiero2011LowThrustMF}.

\begin{equation}
\boldsymbol
  {\vec{f}_{t}} =\sum_{COE} (1-\delta_{COE,COE_{t}})R_{COE}\vec{f}_{COE}
\end{equation}
Where $\vec{f}_{t}$ is the thrust vector in the RIC/RSW frame and  $\delta_{COE,COE_{t}}$ is the Kronecker delta function which is given below;

\begin{equation}
\boldsymbol
    \delta_{COE,COE_{t}}= 
\begin{cases}
    1, & \text COE= COE_{t}\\
    0, & \text COE \neq COE_{t}
\end{cases}
\end{equation}

The DAG targeting algorithm is capable of achieving the desired target but not necessary the resulting path need to be optimal. To generate a better solution, direction weighting factors, $W_{Dir, COE}$ can be included in equation 8  and with consideration that instantaneous COE is not equal to the target COE from equation 9. It can be modified as;

\begin{equation}
\boldsymbol
{\vec{f}_{t}} =\sum_{COE} R_{COE}\vec{f}_{COE} W_{Dir,COE}
\end{equation}

Whereas the $W_{Dir, COE}$  parameter for each COE may be constant or time-varying or maybe the function of some set of the variables in the problem.

In addition to the unit thrust direction, the maneuver efficiencies for each COE are computed along with the maximum rate of the change occurring in each element\cite{doi:10.2514/6.2014-3714}. The equations are mentioned below;

Semi-major axis (a);
\begin{equation}
\boldsymbol
  \theta_{a,max} =0,  \eta_{a} = {\vec{|V|}} \sqrt{\frac{a(1-e)}{\mu(1+e)}}
\end{equation}

\vspace{-3mm}

   Eccentricity (e);
\begin{equation}
\boldsymbol
  \theta_{e,max} =\pi,  \eta_{e} = \frac{1+2e\cos{(\theta)}+\cos^2{(\theta)}}{2(1+e\cos{(\theta)})}
\end{equation}

\vspace{-3mm}

 Inclination (i);
\begin{equation*}
\boldsymbol
  \sin{(\theta_{i}+\omega)}=-e\sin{(\omega)}, 
\end{equation*}
\begin{equation}
\boldsymbol
  {\eta_{i}} = \frac{{\cos{(\omega +\theta)}}}{1+e\cos{(\theta)}} (\sqrt{1-e^2\sin^2{\omega}} -e{\cos{\omega}})
\end{equation}

Ascending Node ($\Omega$);
\begin{equation*}
\boldsymbol
  \cos{(\theta_{\Omega}+\omega)}=-e\cos{(\omega)}, 
\end{equation*}
\begin{equation}
\boldsymbol
  {\eta_{\Omega}} = \frac{{\sin{(\omega +\theta)}}}{1+e\cos{(\theta)}} (\sqrt{1-e^2\cos^2{\omega}} -e{\sin{\omega}})
\end{equation}

Argument of Periapsis ($\omega$);

\vspace{-3mm}
\begin{equation*}
\boldsymbol
  \cos{\theta_{\omega}} =\Bigg[\frac{1-e^2}{2e^3}+\sqrt{(\frac{1-e^2}{8e^3}+\frac{1}{27}})\Bigg]^{1/3}
  \end{equation*}
  \begin{equation*}
\boldsymbol
  -\Bigg[-\frac{1-e^2}{2e^3}+\sqrt{(\frac{1-e^2}{8e^3}+\frac{1}{27}})\Bigg]^{1/3}- \frac{1}{e}
\end{equation*}

\begin{equation}
\boldsymbol
  \eta_{\omega} = \frac{1+\sin^2{\theta}}{1+e\cos{\theta}} * \frac{1+e\cos{\theta_{\omega}}}{1+\sin^2{\theta_{\omega}}}
\end{equation}

For the desired target orbital element, these efficiencies are combined through simple averaging. The DAG law can be used with or without the numerical optimizer, which makes it suitable for the wide range of trajectory simulations for debris cases. 

From the governing equation of Q law and DAG law, we can see that Q-law can cause high computational cost and complexity to simulate as there is no analytical solution while determining the extreme of $\dot{Q}$ which is required at every point of the simulation during the maneuver and de-orbiting operations. Similarly, a convergence of the simulation for maneuver operations to reach nearby to a target location and de-orbiting operation will be time-consuming, which can cause low performance and accuracy of Q law than DAG law. Hence, For our simulation, we will be using highly DAG law for better convergence and accuracy level instead of Q law. Multiple iterative scenarios will be run by considering the DAG for the maneuver, close proximity, and de-orbiting operation in the debris orbit by python package through the STK graphics interface.

\section{Simulation Procedure}

The chase maneuver and the de-orbiting maneuver simulation are done in the STK interface with python run code. To achieve the controlled maneuvers to reach close proximity to debris object and de-orbiting of it, requires all the space environment like perturbation models, debris environment, rendezvous modules, collision avoidance models, etc.
At LEO orbit, the perturbation like an atmospheric drag, the non-elliptical shape of the earth, third body like moon and sun, ocean tides, solar radiation pressure (SRP), earth tides, etc. are encountered and need to be modeled accurately while simulating to achieve higher accuracy level. There is existing orbital propagator like J2, SGP4, HPOP in the literature to the model above effects accurately. Each propagator has its significance to the model of the different perturbation effects. For our case, we had used the HPOP propagator throughout the simulation for better modeling all the perturbation effects encountered by the satellites. The HPOP propagator uses Grace gravity Model(GGM03S) \cite{Tapley2007TheGM}, SW model, EOP model, SOLRESAP model, and SOLFSMY model to provide the Earth gravity field, Space weather data, Earth orientation parameters, Geomagnetic storm indices, and Solar storm indices space environment in the simulation setup. 

The NASA Debris Assessment Software (DAS)  provides us the space debris data set of the tracked debris of the LEO orbit along with its properties like shape, size, orbital 2-line elements, post-mission lifetime, compliance model, etc. of debris which will be helpful while designing the optimal trajectory to chase and de-orbit the PSLV debris. The considered debris is the rocket junk bodies at an approximate altitude of 668 km which its properties can be introduced in simulation by using its 2-line elements with satellite catalog no. of 27160U.

\begin{table}[ht]
\begin{center}
\caption{Initial Debris Chaser and PSLV target orbit.}\label{tab1}%
\begin{tabular}{@{}lll@{}}
\toprule
Parameters & Initial orbit & PSLV target orbit \\

\midrule
Semi-Major axis(km) & 6928.14 & 7046.14 \\ 
Inclination (deg) & 95.0 & 98.3 \\ 
Eccentricity & 0.00472 & 0.00 \\ 
Arg. of Perigee (deg) & 216.9 & Unconstrained \\ 
RAAN & 140.372 & Unconstrained \\
True Anomaly & 120 & Unconstrained \\
Time Epoch & 20 Jan 2022 6:15 & Unconstrained \\
\botrule
\end{tabular}
\end{center}
\end{table}

The debris chaser initial orbit and the PSLV target orbit orbital parameters are shown in table 1. A chase maneuver is to be done to reach close to the PSLV debris target state then after capturing, it will be de-orbiting at an altitude of 250 km.
 The initial states of the debris chaser satellite and the PSLV debris, with reference to the time frame in the ECI frame, are shown in figure 6.
\begin{figure}[ht]
    \centering
    \includegraphics[width =0.5\textwidth,height =2.0in]{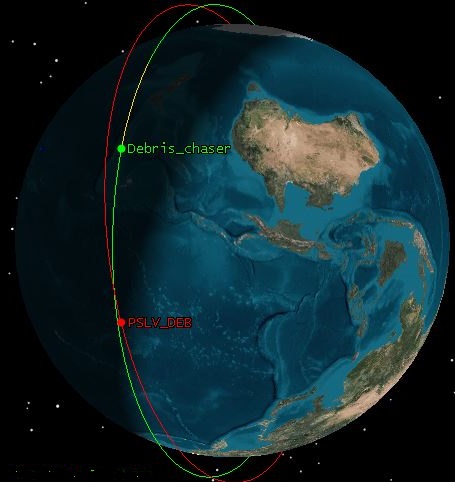}
    \centering
    \caption{Initial states of the Chaser satellite and PSLV Deb.}
    \label{fig:6}
\end{figure}

 From table 1 and figure 6, we can see that, we have to work on the three classical orbital elements of the chaser satellite i.e. semi-major axis and inclination to align with the orbit to the target state and true anomaly to reach nearby the target. This can be done by 3 mission sequence operations. They are;  
 
 a) Hohmann's Orbit transfer operation to reach and align to the orbit of the  PSLV Deb., 
 
 b) V-bar maneuver and close proximity operation to chase the PSLV deb.
 
 c) De-orbiting operation to very low earth orbit.

 \begin{figure}[ht]
    \centering
    \includegraphics[width =0.5\textwidth,height = 3.6in]{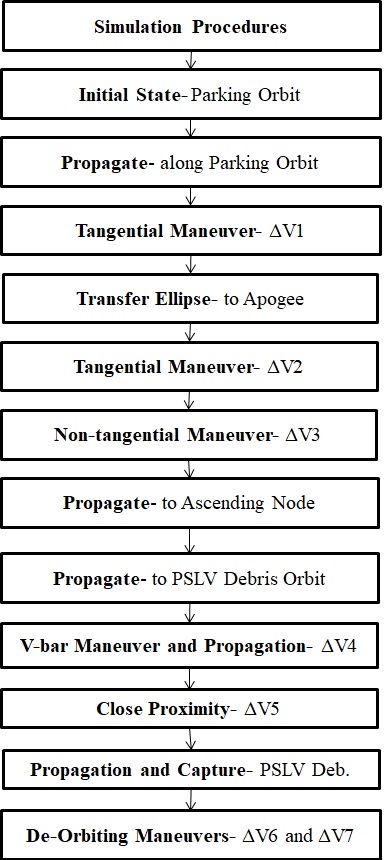}
    \centering
    \caption{ Simulation procedures for debris chaser and de-orbiting operation.}
    \label{fig:7}
\end{figure}

The simulation procedures are shown in figure 7. Figure 7 represents the flow of the process from the orbit transfer to the de-orbiting operation whereas the capture techniques are not presented in this paper.

The snapshots of the v-bar approach of the chaser satellite to chase the debris are shown in figure 8. It forms the circumnavigated trajectory to reach close to the debris objects. The closest in-track separation between the debris chaser satellite and PSLV debris is about 1 m. The close proximity snapshot between the chaser and debris satellite is shown in figure 9. Once reaching too close, the robotic manipulator will be used for grasping and ceasing the random motion of the PSLV debris.
 
  \begin{figure}[ht]
   \centering
     \begin{minipage}{.48\textwidth}
      \centering
      \includegraphics[width =0.95\textwidth,height =2.0in]{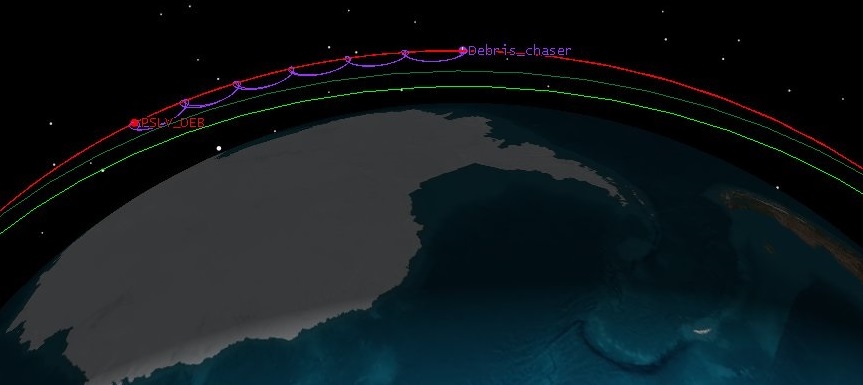}
      \centering
      \caption{ V-bar rendezvous trajectory formation to chase PSLV Deb.}
       \label{fig:8}
       \end{minipage}
    \begin{minipage}{.48\textwidth}
    \centering
      \includegraphics[width =0.95\textwidth,height =2.0in]{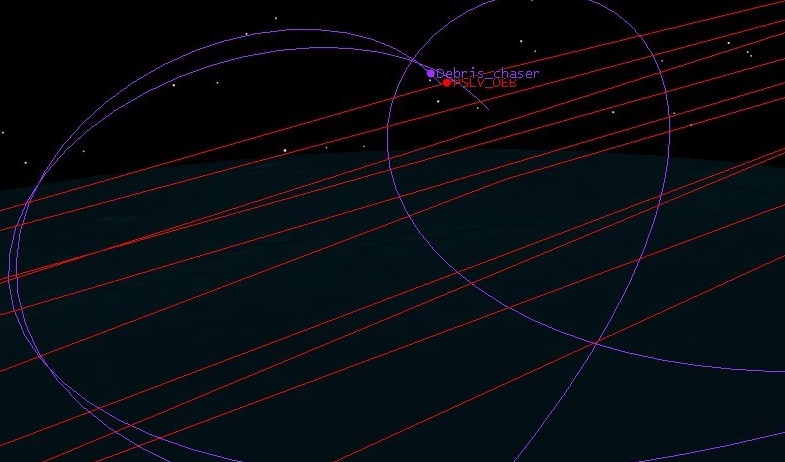}
     \centering
      \caption{Close proximity between the Chaser satellite and the PSLV Deb.}
      \label{fig:9}
     \end{minipage}
\end{figure}

\section{Results and Discussion}

The simulation was run by using the DAG law for chase maneuver(V-bar approach and close proximity) and the de-orbiting operation. To achieve these operations it almost took  18 hours of the operation which include 2 hours for the capturing process, once reaching close enough to PSLV debris. The overall delta requirement for these operations as followed by the simulation procedure is shown in table 2.

\begin{table}[ht]
\begin{center}
\caption{Over all $\Delta V$ requirement tabulation during the operations.}\label{tab2}%
\begin{tabular}{@{}llllll@{}}
\toprule
 SN.   & Debris Chaser Segment  &  Start time & $V_{x}(m/s )$  & $V_{y}(m/s )$  & $V_{z}(m/s )$\\
\midrule
1. & Tangent-$\Delta$V1 & 20 Jan 2022 07:45:00 & 31.9573 & 0  & 0  \\ 
2. & Tangent-$\Delta$V2  & 20 Jan 2022 08:25:08 & 31.8226 & 0  & 0  \\ 
3. & Non-tangent-$\Delta$V3  & 20 Jan 2022 09:34:47 & -13.9376 & 433.119  & 33.3713 \\ 
4. & Non-tangent-$\Delta$V4  & 20 Jan 2022 10:44:45 & 75.1297 & -32.1772 & -12.7527 \\ 
5. & Non-tangent-$\Delta$V5  & 20 Jan 2022 20:44:45 & 42.5346 & 30.9693  & -3.31569 \\ 
6. & Tangent-$\Delta$V6 & 20 Jan 2022 22:18:40 & -115.849 & 0  & 0  \\ 
7. & Tangent-$\Delta$V7  & 20 Jan 2022 23:07:48 & -117.634  & 0  & 0  \\ 
\botrule
\end{tabular}
\end{center}
\end{table}

\vspace{-5mm}
Table 2 represents steps 1-3 which is the initial Hohmann's transfer from low earth orbit to debris orbit which is executed by 2 tangent burns and one non-tangent burn for inclination change. Once reaching debris orbit, a couple of non-tangent burns (i.e. steps 4-5) is made for the controlled v-bar rendezvous approach and close proximity operation. After that, the space manipulators are used to capture the non-cooperative PSLV debris and cease their motion, and orient with debris chaser satellite. Once it becomes a single body, again two tangent burn (i.e.step 6-7) is made for de-orbiting operation from debris orbit to very low earth orbit. i.e. 250km.

Once, the controlled operations are executed, the variation of the orbital elements, velocity vector field, and RIC frame parameters was sketched for all the simulation time frames of PSLV debris and Chaser satellites. And these sketched parameters were analyzed technically.

\begin{figure}[ht]
    \centering
    \includegraphics[width =0.98\textwidth,height =2.0in]{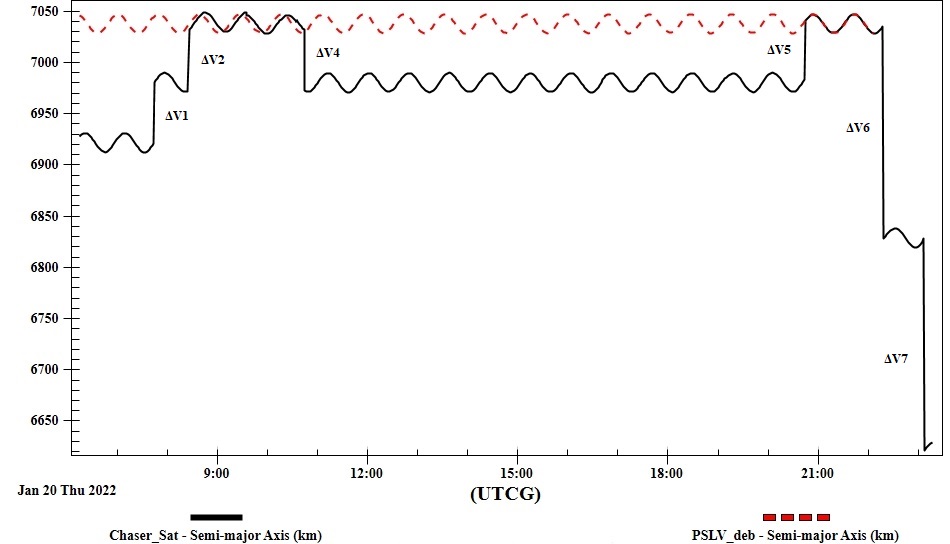}
    \centering
    \caption{Variation of Semi-Major Axis (a) across the time.}
    \centering
    \label{fig:10}
\end{figure}

\vspace{-4mm}
The graphical representation of the semi-major axis with respect to time is shown in figure 10. From it, an altitude shift is taking place from low earth orbit to debris orbit by means of $\Delta$V1 and $\Delta$V2 tangent burn. And the sinusoidal motion is due to the perturbation effects considered in the LEO environment. The $\Delta$V1 and $\Delta$V2 cause the transfer is not a straight line, which is actually inclined and impulsive maneuver. The maneuvers are taking just few a minutes while making a transfer at constant thrust. And the $\Delta$V3  is required for the inclination change once reaching that height

After the execution of transfer operation, as the orbit gets stabilized after 2 orbits, then we go for v-bar rendezvous and proximity operation. From figure 10, we can see again, $\Delta$V4 non-tangent burn is done at 10:44 am to initiate the V bar chase approach to reach near to debris and form the circumnavigate motion up to 10 hr period, causing the periodic height change. Once reaching close to the debris location, the $\Delta$V5 burn is done to stabilize and align the radial and cross-track of the chaser satellite along with the debris satellite within track separation of 1m. This process is simulated in a controlled maneuver way with the help of DAG law and provides stabilized chase towards the target debris.

\begin{figure}[ht]
\centering
     \begin{minipage}{.49\textwidth}
    \centering
    \includegraphics[width =0.98\textwidth,height =1.8in]{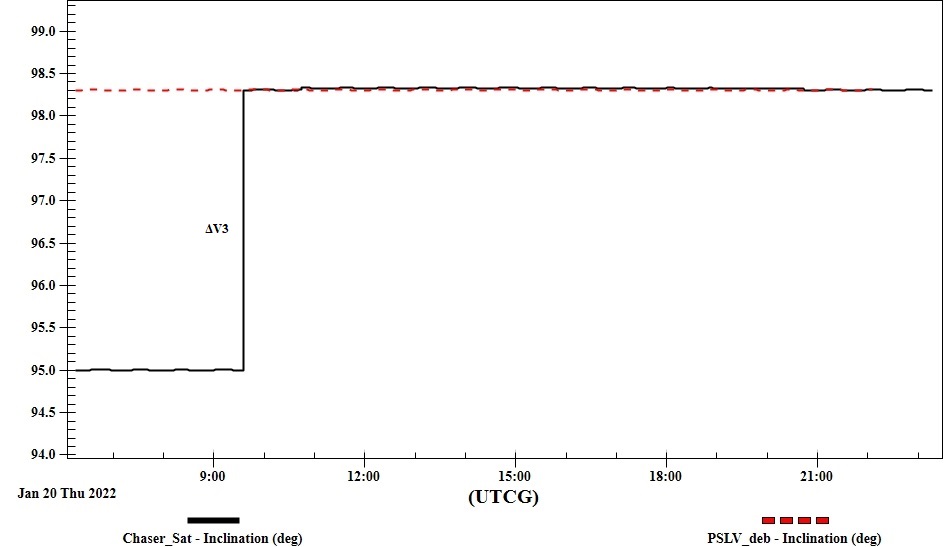}
    \centering
    \caption{ Variation of inclination (i) across time. }
    \label{fig:11}
    \end{minipage}
    \centering
     \begin{minipage}{.49\textwidth}
    \centering
    \includegraphics[width =0.98\textwidth,height =1.8 in]{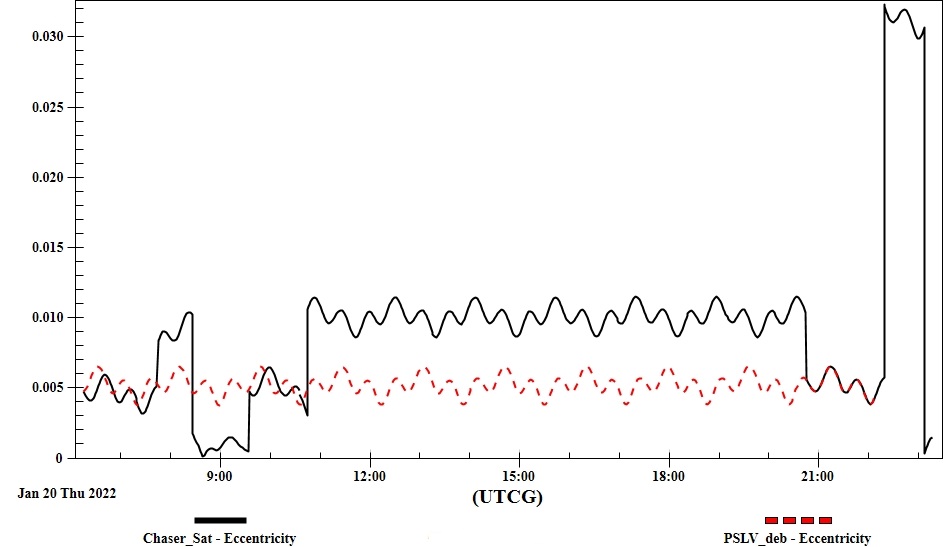}
    \centering
    \caption{ Variation of the eccentricity (e) across time.  }
    \label{fig:12}
    \end{minipage}
\end{figure}

\vspace{-5mm}
The graphical representation of the variation of inclination and the eccentricity of simulated simulation with respect to time is shown in figures 11 and 12. we can see that the chaser satellite was initially at 95 degrees and remain constant until the non-tangential  $\Delta$V3 is applied to change inclination to 98.3 degrees as of the debris chaser orbit. As being an impulsive maneuver, this non-tangent transfer to make orbital plane shift also take less than a minute time.

The change that occurred in the semi-major axis and the inclination by the $\Delta$V budget causes the change in the other orbital parameters in the same fashion.
From figure 12, we can see the variation of the eccentricity with respect to the time, as both debris chaser and the PSLV Deb. orbit is circular, so most of the eccentricity is value is close to zero. we can see that the slight change in the eccentricity caused during the $\Delta$V applied to section and maximum change that occurred is about 0.03.

 The schematic diagram of the variation of the RAAN with respect to the simulation time is shown in figure 22. From figure 22, we can see that there is much less change occurred in the initial and the final value of the RAAN of the chaser satellite during the whole operation.

\begin{figure}[ht]
\centering
     \begin{minipage}{.49\textwidth}
    \centering
    \includegraphics[width =0.98\textwidth,height =1.8in]{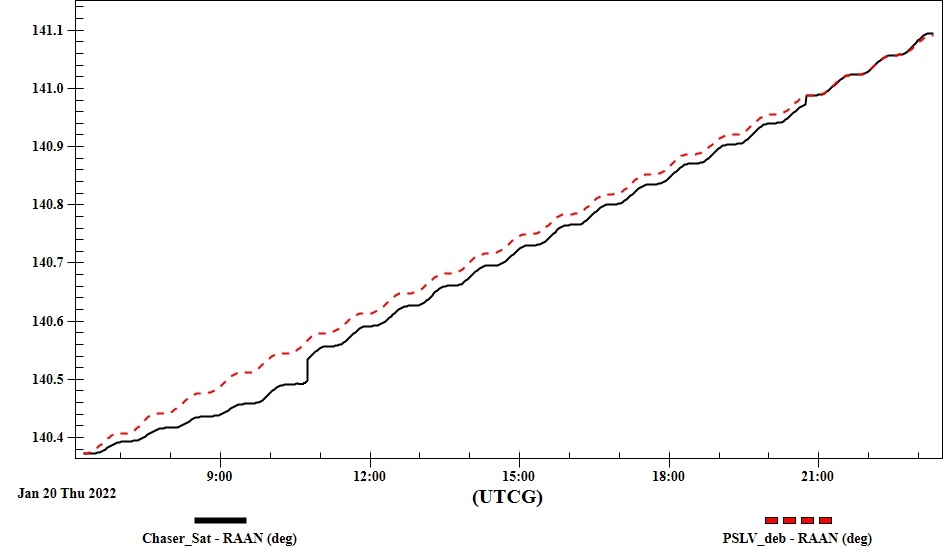}
    \centering
    \caption{ Variation of the RAAN ($\Omega$) across time. }
    \label{fig:13}
    \end{minipage}
    \centering
     \begin{minipage}{.49\textwidth}
    \centering
    \includegraphics[width =0.98\textwidth,height =1.8in]{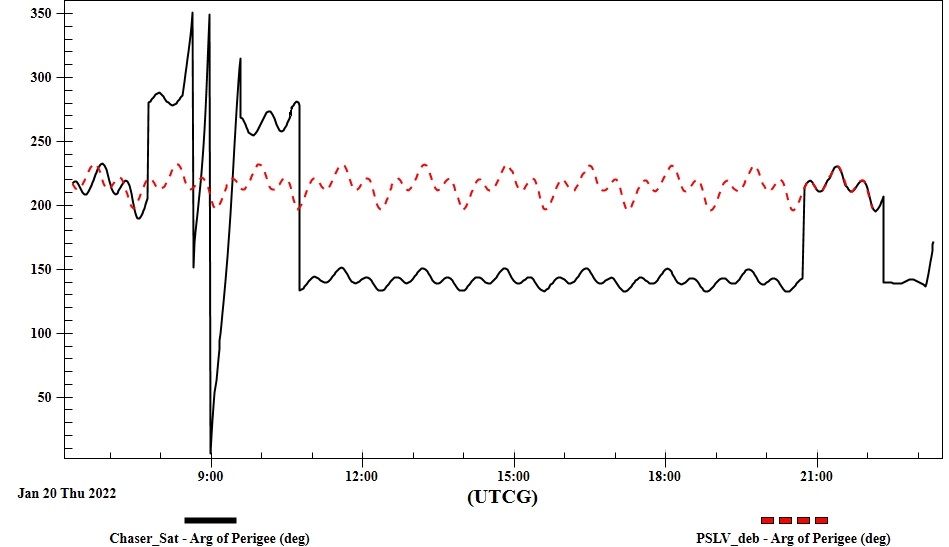}
    \centering
    \caption{ Variation of the Argument of Periapsis ($\omega$) across  time. }
    \label{fig:14}
    \end{minipage}
\end{figure}

\vspace{-5mm}
The graphical representation of a variation of RAAN and Argument of Periapsis with respect to time is shown in figures 13  and 14.
The orbital plane was mostly defined by the inclination and RAAN and we found that there are 3.3 degrees of an orbital shift to align to debris inclination. This shift causes a slight change in the RAAN of the chaser satellite and continuous convergence takes place during rendezvous operation and de-orbiting operation and the final RAAN of debris chaser matches with the PSLV debris RAAN. We can see the maximum variation that occurred during the $\Delta$V operations time frame as caused in the semi-major axis and the eccentricity. And maximum change occurs during the  $\Delta$V3 time  frame where inclination change had caused the peak change in the argument of periapsis. And after the end of rendezvous operations, the argument of periapsis matches the argument of periapsis of PSLV debris and de-orbiting variation takes place at the end. 

\begin{figure}[ht]
    \centering
    \includegraphics[width =0.98\textwidth,height =1.8in]{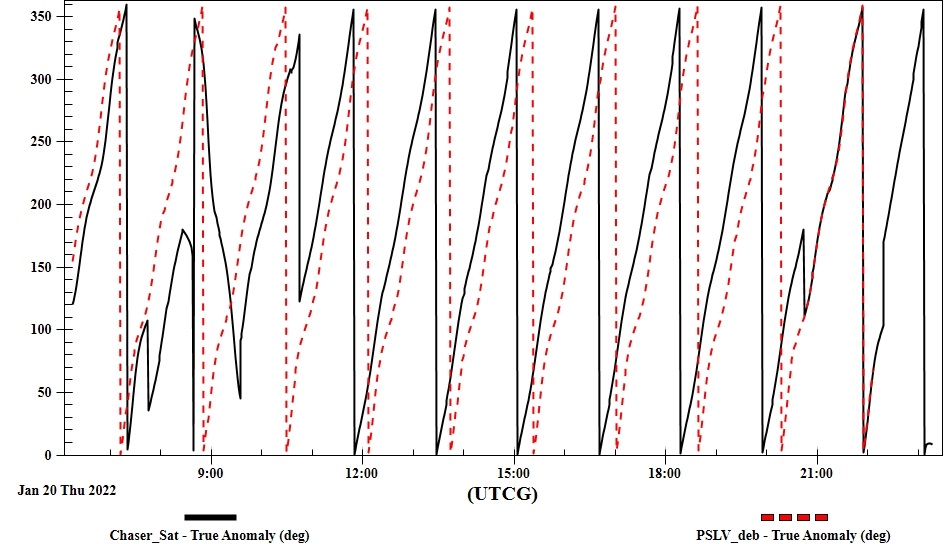}
    \centering
    \caption{ Variation of the True Anomaly($\nu$) across time. }
    \label{fig:15}
\end{figure}

 The position of the satellite in an orbit is mostly determined by the true anomaly. It varies periodically from 0-360 degrees in an orbit. This parameter will help us to find the true anomaly separation between the debris chaser satellite and PSLV debris and help us to do rendezvous operations accordingly. The graphical representation of the true anomaly is shown in figure 15. At maneuver location, there is a slight shift taking place and at the end of close proximity operation, it will be in the same time with PSLV debris true anomaly and de-orbiting take place at the end which causes a slight shift in a true anomaly.
 
 Similarly, we had generated the result of the velocities parameter of debris chaser satellite and PSLV debris in all x,y, and z-axis. The schematic diagram is shown in figures 16, 17, and 18.
 
 \begin{figure}[ht]
\centering
     \begin{minipage}{.49\textwidth}
    \centering
    \includegraphics[width =0.98\textwidth,height =1.8in]{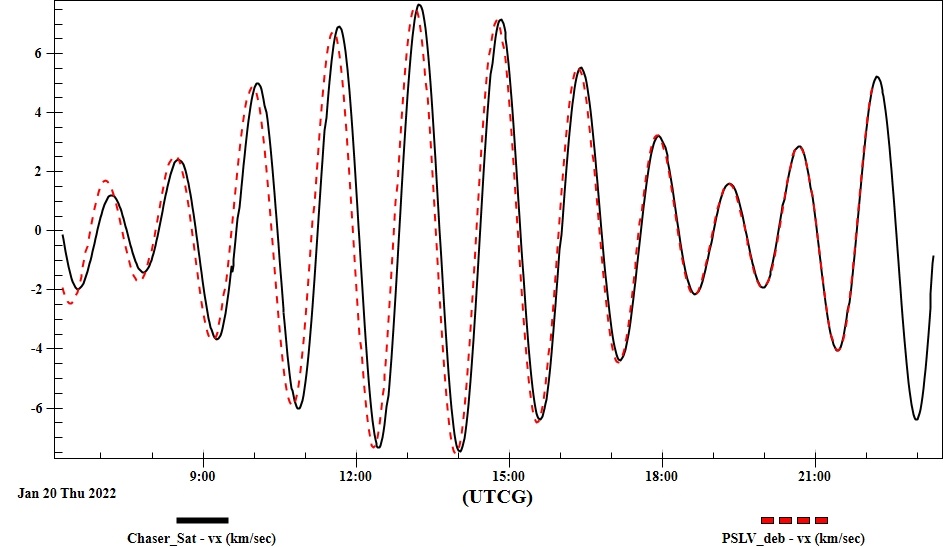}
    \centering
    \caption{ Variation of Vx across time. }
    \label{fig:16}
     \end{minipage}
\centering
     \begin{minipage}{.49\textwidth}
    \centering
    \includegraphics[width =0.98\textwidth,height =1.8in]{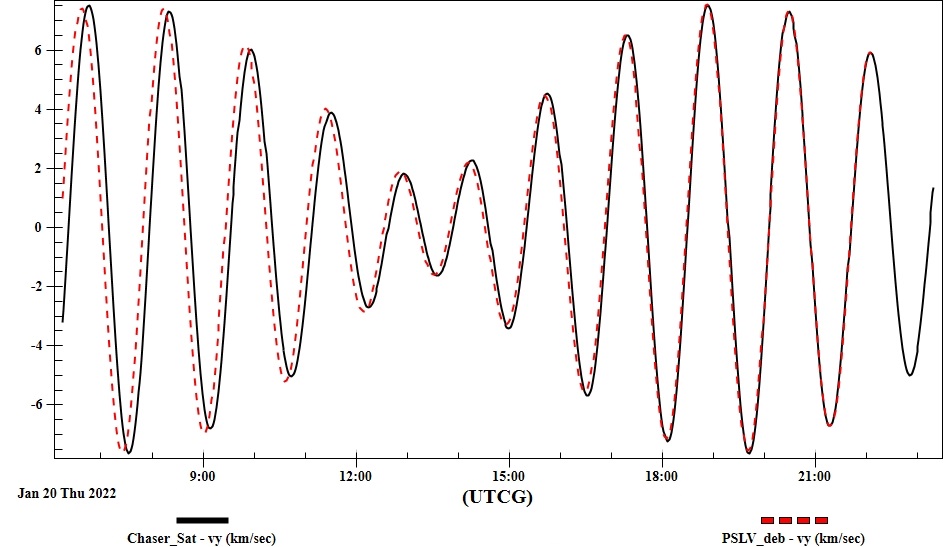}
    \centering
    \caption{ Variation of Vy across time. }
    \label{fig:17}
     \end{minipage}
\end{figure}
 
 \vspace{-8mm}
 \begin{figure}[htp]
    \centering
    \includegraphics[width =0.98\textwidth,height =1.8in]{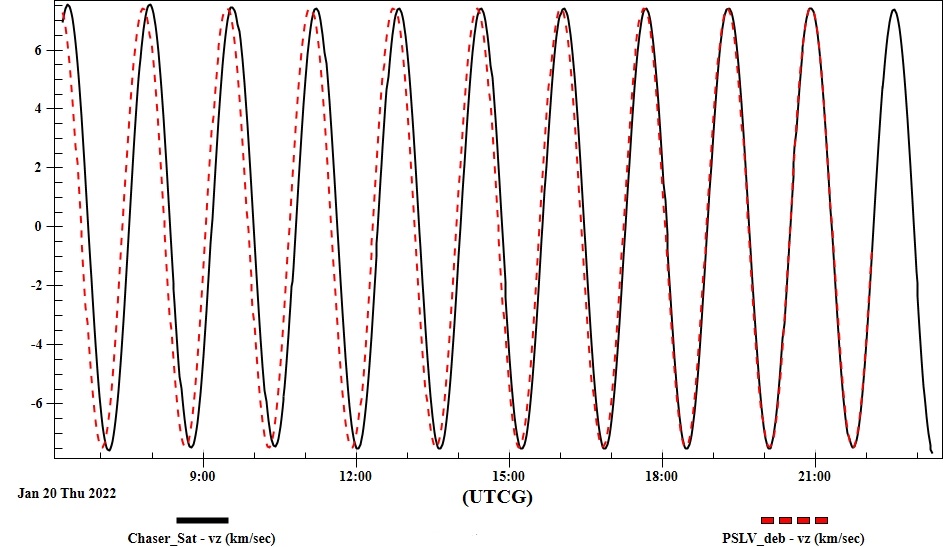}
    \centering
    \caption{ Variation of Vz across time. }
    \label{fig:18}
\end{figure}
 
 Since, all the above results are shown in orbital elements and velocity vector, which represents the shift and how we are controlling each orbital element at a time to reach nearby the PSLV debris location and controlled de-orbiting operation.
But, RIC frame parameters become important to analysed to know where our satellite is located in orbit with respect to PSLV debris. To do so, we had graphically represented the variation of radial, in-track, and cross-track components with reference to debris target with respect to time. These schematic diagrams are shown in figures 19, 20, and 21.
 
\begin{figure}[ht]
\centering
     \begin{minipage}{.49\textwidth}
    \centering
    \includegraphics[width =0.98\textwidth,height =1.8in]{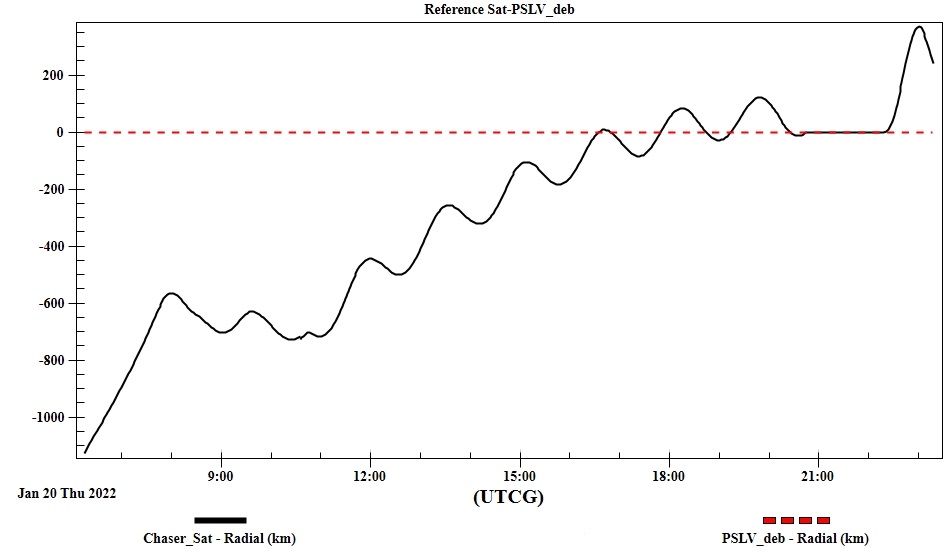}
    \centering
    \caption{ Variation of the radial component across time. }
    \label{fig:19}
     \end{minipage}
\centering
     \begin{minipage}{.49\textwidth}
    \centering
    \includegraphics[width =0.98\textwidth,height =1.8in]{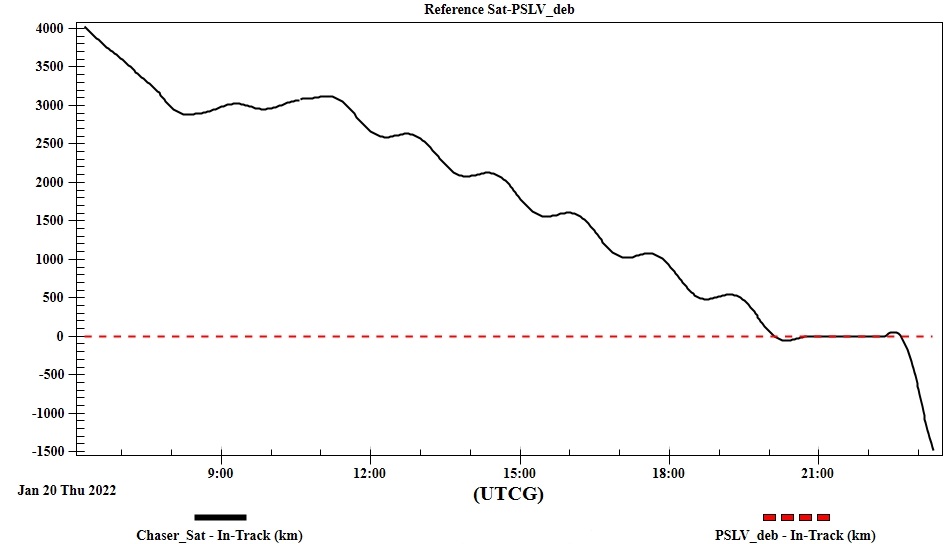}
    \centering
    \caption{ Variation of the in-track component across time. }
    \label{fig:20}
     \end{minipage}
\end{figure}

 From figure 19 and 20, we can find that the gap in radial distance and in-track component between the chaser satellite and the PSLV Deb is decreasing continuously and at the end of the rendezvous operation, the radial and in-track is almost in the same line of the debris orbit with in-track separation of 1m displacement. Once the capturing operation is done at close proximity, the de-orbiting operation is done where the gap is continuously increasing in both radial and in track components.

\begin{figure}[ht]
    \centering
    \includegraphics[width =0.98\textwidth,height =1.8in]{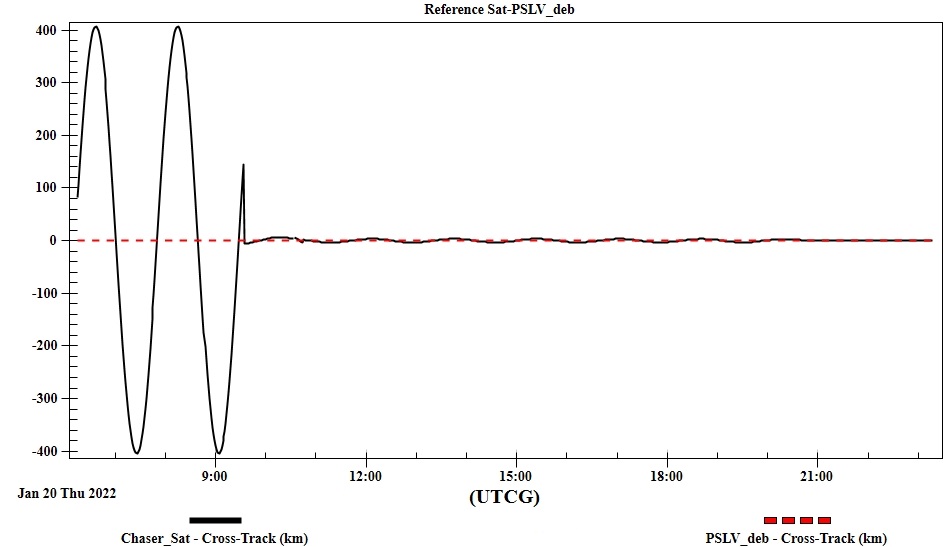}
    \centering
    \caption{ Variation of the cross-track component across time. }
    \label{fig:21}
\end{figure}

Similarly, the cross-track components of the chaser satellite with reference to the PSLV Deb along with the time are sketched in figure 21. The cross-track component varies sinusoidally during the orbit transfer operation and at close proximity, the cross track aligns with the PSLV debris cross-track component. In de-orbiting operation, the cross-track remains the same as PSLV debris as it acts as a combined body.

The tabulation of the RIC components with desired, achieved and the difference was shown in table 3.

\begin{table}[ht]
\begin{center}
\caption{Tabulation of the desired, achieved and difference value in RIC frame with Reference to PSLV deb. at close proximity operation.}\label{tab3}%
\begin{tabular}{@{}llll@{}}
\toprule
RIC frame & Desired(Km) & Achieved(Km)  & Difference(in Km)\\
\midrule
Radial & 0 & -1.30732e-7   & -1.30732e-7\\ 
In-track & 0.001 & 0.00100145  & 1.4538e-6 \\ 
Cross-track & 0 & -3.9958e-5 & -3.9958e-5 \\
\botrule
\end{tabular}
\end{center}
\end{table}

From table 3, we can see the RIC frame components during the close proximity operation with desired, achieved, and their difference values.
During close proximity operation, the radial component will be in same as to PSLV debris so the desired value is approximate 0 km and from the iterative operation, we found we had achieved -1.30732e-7 km separation which is desirable. Similarly, the in-track separation needs to be at 1m between the debris chase and PSLV debris but we achieve the approximate same value. and in cross-track, we achieved -3.9958e-5 separation which is highly acceptable. These differences can be corrected by using the ADCS system of satellites with efficient sensors.

From the results section, we can see that we had graphically represented and analyzed all the orbital elements, orbital velocities along with the RIC frame components variation with respect to the time of all operations. The delta-V budget required for the successful operation of this simulation and close proximity RIC components achievement were tabulated and the accuracy level was analyzed.

\section{Conclusion}

A low thrust controlled maneuver to chase and de-orbit the space debris was designed by using the debris chaser satellite with dual manipulators. The paper describes the debris chaser mission whose main function will be to chase, capture and de-orbit the PSLV debris of polar orbit.
The designed satellites are made based on the CubeSat standards with 12U form factor and components were selected based on the commercial products available in the market. A debris chaser system architecture was presented with the block diagram with all the important components mentioned on it. And it was found that ADCS sensors will play the most vital role while executing the chase and de-orbiting operations.

For our case, we had to use the basic Hohmann's transfer concept for orbit transfer and the V-bar approach for the chase maneuver along with the DAG guidance algorithm for the controlled process. Most of the orbit transfer, chase maneuver, and de-orbiting operations were executed in the STK interface with the python pipeline.  We had also discussed the use and importance of the DAG law in our simulation over the Q-law. In a later section, we had explained the simulation procedures and their environment setup in our simulation along with its process for the execution of the operations.

Once a simulation is completed, multiple datasets like the orbital element, velocity vector and RIC frame components are extracted and plotted against the time. All the delta-V budget was tabulated with their start time which is required while executing the simulation operation. It was found that the use of DAG law has provided smooth control operation for chase and de-orbiting maneuvers. All the sketched graphs provide the variation of parameters with respect to the time and their importance during the execution of the operations shows stable results while chasing and de-orbiting the space debris.

The sketched RIC frame components of the chaser satellite by taking reference of the PSLV debris provide the radial, in-track, and cross-track variation with respect to debris. It was found that the radial and cross-track separation during the close proximity operation shows the desirable results which were achieved successfully. Similarly, an in-track separation between the debris chaser satellite and the PSLV debris is desired to be 1m which was achieved successfully. Hence, we can assure that, for the long-term simulation of the rendezvous, close proximity maneuver, and de-orbiting operation, the optimal DAG law is more effective in simulation as it can handle the singularity behavior of the orbital elements caused due to adjustment of one or more elements more efficiently. And the use of RCS thusters for our operations has successfully provided the required thrust based on the delta-V budget in all directions.
The next work will be of grasping the non-cooperative target debris and ceasing the motion of the debris by the use of thruster and ADCS system.

\bibliography{sn-bibliography}


\end{document}